\documentclass[iicol,sn-apa]{sn-jnl}% Default with double column layout

\usepackage{graphicx}%
\usepackage{multirow}%
\usepackage{amsmath,amssymb,amsfonts}%
\usepackage{amsthm}%
\usepackage{mathrsfs}%
\usepackage[title]{appendix}%
\usepackage{xcolor}%
\usepackage{textcomp}%
\usepackage{manyfoot}%
\usepackage{booktabs}%
\usepackage{algorithm}%
\usepackage{algorithmicx}%
\usepackage{algpseudocode}%
\usepackage{listings}%

\usepackage{subcaption}
\begin{document}

\title[Article Title]{Pothole Detection and Recognition based on Transfer Learning}

\author[1]{\fnm{Mang} \sur{Hu}}\email{humang@cug.edu.cn}

\author[1]{\fnm{Qianqian} \sur{Xia}}\email{xiaqianqian2024@163.com}

% \author*[1,2]{\fnm{Linquan} \sur{Yang}}\email{yanglq@cug.edu.cn}

\affil*[1]{\orgdiv{School of Computer Science}, \orgname{China University
of Geosciences}, \orgaddress{\city{Wuhan}, \postcode{430078}, \country{China}}}

% \affil[2]{\orgdiv{Hubei Key Laboratory of Intelligent Geo-Information
% Processing}, \orgname{China University
% of Geosciences}, \orgaddress{\city{Wuhan}, \postcode{430078}, \country{China}}}

\abstract{With the rapid development of computer vision and machine learning, automated methods for pothole detection and recognition based on image and video data have received significant attention. It is of great significance for social development to conduct an in-depth analysis of road images through feature extraction, thereby achieving automatic identification of the pothole condition in new images. Consequently, this is the main issue addressed in this study. Based on preprocessing techniques such as standardization, normalization, and data augmentation applied to the collected raw dataset, we continuously improved the network model based on experimental results. Ultimately, we constructed a deep learning feature extraction network ResNet50-EfficientNet-RegNet model based on transfer learning. This model exhibits high classification accuracy and computational efficiency. In terms of model evaluation, this study employed a comparative evaluation approach by comparing the performance of the proposed transfer learning model with other models, including Random Forest, MLP, SVM, and LightGBM. The comparison analysis was conducted based on metrics such as Accuracy, Recall, Precision, F1-score, and FPS, to assess the classification performance of the transfer learning model proposed in this paper. The results demonstrate that our model exhibits high performance in terms of recognition speed and accuracy, surpassing the performance of other models. Through careful parameter selection and model optimization, our transfer learning model achieved a classification accuracy of 97.78\% (88/90) on the initial set of 90 test samples and 98.89\% (890/900) on the expanded test set.}

% \pacs[JEL Classification]{D8, H51}


\maketitle

\section{Introduction}\label{sec1}
Pothole detection~\citep{tsai2018pothole} refers to the utilization of computer vision and image processing techniques to analyze road images and identify the presence and location of potholes on the road. The objective of this technology~\citep{ouma2017pothole} is to provide accurate road condition information for timely maintenance and repair, thereby enhancing traffic safety and driving comfort. Its significance extends beyond the realm of transportation and plays a crucial role in geological exploration, aerospace science, and natural disaster management.

In recent years, with the continuous development of intelligent hardware and deep learning~\citep{kumar2020modern}, several recognition, detection, and segmentation problems in computer vision have been effectively addressed. Compared to traditional machine learning methods, deep learning eliminates the need for manual feature engineering. With the use of deep neural networks, deep learning can automatically adapt and extract the required features for detection. Additionally, deep learning generally outperforms traditional image recognition methods in terms of detection accuracy.

In recent times, deep learning techniques have brought new solutions to pothole detection. Through computer vision technology, we can extract key features such as pothole contours, textures, and shapes from pothole images. These features can be transformed into more easily classifiable representations, enabling the use of pattern recognition and classification algorithms to determine the presence of potholes. The application of these technologies is not only significant in the field of transportation but also plays an important role in research and practical applications in other domains. Currently, the main issues in road pothole detection technology lie in the accuracy and robustness of detecting potholes within the drivable areas under different backgrounds. Most existing algorithms are not suitable for complex and diverse road scenarios, and some algorithms fail to strike a balance between accuracy, real-time performance, and robustness. Consequently, extensive research efforts are still required to advance road pothole detection technology. In light of these circumstances, this paper proposes a deep learning approach based on transfer learning, utilizing ResNet50, EfficientNet, and RegNet convolutional neural networks as the foundation, and further enhancing them to precisely locate potholes on the road surface. Finally, the feasibility and effectiveness of this method are verified through experiments.

In conclusion, this paper makes three main contributions to the road pothole detection task:
\begin{itemize}
\item We propose a deep learning approach based on transfer learning, utilizing ResNet50, EfficientNet, and RegNet convolutional neural networks as the foundation. These networks demonstrate high accuracy on the IMAGENET dataset while maintaining relatively lower parameter count and FLOPS. We replace the original fully connected layers at the end of the networks with a unified customized 5120x2 fully connected layer, enabling more comprehensive mapping of extracted features to the sample's label space.
\item We improve the loss function by incorporating both the commonly used cross-entropy loss function in classification tasks and the L1 loss function. This enhancement enhances the robustness, generalizability, and feature selection of the model, resulting in further optimization of the model performance.
\item Experimental results demonstrate that our proposed method outperforms traditional methods in terms of both recognition accuracy and speed.
\end{itemize}

\section{Related Work}\label{sec2}
Road pothole detection research worldwide has explored various approaches based on computer vision. These approaches can be broadly categorized into three types: algorithms that directly process road images using classical 2D image processing techniques such as enhancement, compression, transformation, and segmentation; algorithms that utilize 3D road point cloud modeling and segmentation to match specific geometric models to observed point clouds, enabling the identification of potholes; and methods based on deep learning, which employ image classification, object detection, or semantic segmentation algorithms to detect road potholes by training neural networks on labeled data. These different approaches reflect the diverse strategies employed to address the challenge of road pothole detection, with ongoing efforts to enhance accuracy and efficiency in this field.

Classic 2D image processing-based methods for road pothole detection typically involve four main processes: image preprocessing, image segmentation, damage area extraction, and post-processing of detection results. FAN~\citep{fan2019pothole} utilized the OTSU thresholding method to segment transformed disparity images for road pothole detection. Additionally, a simple linear iterative~\citep{fan2021rethinking}  clustering algorithm was employed to group the transformed disparities into superpixel sets, enabling the detection of road potholes by identifying these superpixels. This approach achieved higher efficiency and accuracy in pothole detection. However, despite nearly 20 years of research history in road pothole detection using classical 2D image processing methods, these techniques are built upon earlier technology and are susceptible to environmental factors such as lighting and weather conditions. In contrast, 3D technology and deep learning algorithms have significantly overcome these limitations, offering more robust and advanced solutions in road pothole detection.

The overall process of road pothole detection based on 3D point cloud modeling and segmentation can be divided into two stages. Firstly, the observed 3D road point cloud data is interpolated to obtain an explicit geometric model, which can be a plane or a quadratic surface. Next, the observed point cloud is compared with the established model to separate the pothole regions from the 3D road point cloud. For instance, ZHANG~\citep{zhang2013advanced} utilized the least squares fitting method to interpolate the observed 3D point cloud onto a specific plane. By searching for 3D points below the fitted surface, potential road potholes were roughly detected. Subsequently, the detection results were further refined using the K-means clustering algorithm and region growing algorithm. Similarly, RAVI~\citep{ravi2020highway} designed a road pothole detection system based on lidar, categorizing the road points into damaged and undamaged classes by comparing the distances between the 3D road points and the best-fitted 3D road surface. However, it should be noted that road surfaces are not always perfectly flat in reality, which can pose challenges for 3D point cloud modeling and segmentation methods. Additionally, if the goal is only to identify and locate potholes without the need for geometric details such as size and depth, it may not be necessary to employ laser devices for extracting 3D point clouds.

With the rapid development of object detection algorithms, there are now more choices available for pothole detection, in addition to traditional methods. Object detection algorithms can be categorized into single-stage and two-stage detection algorithms. Two-stage detection algorithms first extract candidate regions from the image and then refine the detection results based on these candidate regions. They usually achieve higher detection accuracy but are slower in terms of processing speed. For example, the pioneering work in object detection, RCNN~\citep{girshick2014rich}, achieved higher accuracy compared to traditional methods but had a complex and time-consuming training process and struggled to find the global optimal solution. Subsequently, Girshick~\citep{girshick2015fast} improved upon RCNN by incorporating the feature pyramid concept and introduced Fast R-CNN, which introduced a new multi-task loss function that reduced the computational resources required for object detection. The Faster R-CNN~\citep{ren2015faster} model enabled end-to-end training, improving the speed of algorithm training and execution. However, the fully connected layers in Faster R-CNN could lead to information loss, affecting the accuracy of the algorithm. Later, Mask R-CNN~\citep{he2017mask} was proposed by adding parallel mask branches to Faster R-CNN, addressing the problem of simultaneous object localization, classification, and segmentation. However, its real-time performance still falls short of ideal expectations.

Based on the above situation, this paper constructs a deep learning feature extraction network based on transfer learning to extract road image features. We have chosen a fusion approach that combines three networks: ResNet50, EfficientNet, and RegNet, known as the ResNet50-EfficientNet-RegNet model, to extract image features. Furthermore, we replace the original fully connected layer at the end of the network with a customized 5120*2 fully connected layer to output the classification results of the images, while also improving the loss function. Consequently, this research achieves road pothole detection. In comparison with traditional detection algorithms, it enhances the accuracy, speed, and generalization ability of the detection.

\section{Method}\label{sec3}
In terms of model building, this study employed a transfer learning approach to classify pothole images by utilizing the features learned from a pre-trained model designed for general image recognition tasks. Transfer learning~\citep{pan2009survey} is a deep learning technique that transfers knowledge obtained from one task to another related task, thereby accelerating and improving the learning process. In the context of deep learning, transfer learning is often based on pre-trained neural network models. The model development in this study incorporated a fusion of three networks, namely ResNet50, EfficientNet, and RegNet, through transfer learning. This fusion resulted in the ResNet50-EfficientNet-RegNet model, which was specifically designed for pothole detection and recognition in road images.

Therefore, we have chosen these three networks as the core of our model for several reasons. Firstly, they have demonstrated the highest accuracy on the IMAGENET dataset, which is widely regarded as a benchmark in computer vision tasks. Secondly, these networks exhibit relatively lower parameter counts and FLOPS (floating-point operations per second) compared to other alternatives. The specific data illustrating these characteristics are presented in the following table:

\begin{table}[h]
\centering
\caption{\small{Comparison of models}}
\begin{tabular}{l c c c c}
\toprule
Model& Acc@1 & Acc@5 & Params & GFLOPS \\
\midrule
GoogLeNet & 69.778 & 89.53 & 6.6M & 1.5 \\
VGG19 & 72.376 & 90.876 & 143.7M & 19.63 \\
MobileNet & 75.274 & 92.566 & 5.5M & 0.22 \\
ShuffleNet & 76.23 & 93.006 & 7.4M & 0.58 \\
DenseNet201 & 76.896 & 93.37 & 20.0M & 4.29 \\
ViT & 76.972 & 93.07 & 306.5M & 15.38 \\
ResNeXt101 & 79.312 & 94.526 & 88.8M & 16.41 \\
ResNet50 & 80.858 & 95.434 & 25.6M & 4.09 \\
EfficientNet & 84.228 & 96.878 & 21.5M & 8.37 \\
RegNet & 82.828 & 96.33 & 39.4M & 8.47 \\
\bottomrule
\end{tabular}
\label{tab:11}
\end{table}

\subsection{Model Architecture}
\subsubsection{ResNet50}
ResNet50~\citep{he2016deep} is a specific model within the ResNet series, which is an improved version of the original ResNet model. The number "50" in ResNet50 represents the total number of layers, including convolutional layers, pooling layers, fully connected layers, etc. Compared to shallower ResNet models, ResNet50 has a deeper network architecture. The following are key structures present in ResNet50 that contribute to its exceptional performance in computer vision tasks such as image classification, object detection, and image segmentation.

1)	Residual Connections: ResNet50 introduces residual connections, which address the issues of vanishing and exploding gradients by establishing direct connections across layers. This connection allows information to flow directly to subsequent layers, making the network easier to train. The use of residual connections helps alleviate the challenges associated with increasing network depth while improving the accuracy of the network.

2)	Stacking Convolutional and Pooling Layers: In ResNet50, a series of convolutional and pooling layers are stacked at the beginning of the network to extract low-level features from the input image. The stacking of these layers gradually enlarges the receptive field and enhances the abstraction capability of the features.

3)	Bottleneck Structure: The bottleneck structure in ResNet50 consists of 1x1, 3x3, and 1x1 convolutional layers. This structure reduces the computational complexity of the model while providing more powerful feature representation. The bottleneck structure enables more efficient learning of complex features in images, allowing the network to be deeper and possess better expressive power.

4)	Global Average Pooling: After the last convolutional layer in ResNet50, a global average pooling layer is applied to the feature maps to reduce their spatial dimensions. This pooling operation reduces the spatial dimensions of the feature maps to 1 while preserving the feature channels, enabling the network to better capture the global information of the overall image.

Through the combination and design of these architectural components, ResNet50 effectively addresses the gradient problem in deep networks and possesses stronger feature extraction and abstraction capabilities, leading to outstanding performance in classification tasks.

\subsubsection{EfficientNet}
EfficientNet~\citep{tan2021efficientnetv2} is a highly efficient and powerful Convolutional Neural Network (CNN) architecture proposed by the Google research team. It achieves better performance with relatively fewer computational resources by employing a unified scaling strategy across network depth, width, and resolution. The design philosophy of EfficientNet is to improve the accuracy and efficiency of the model through modifications to the network structure and parameter adjustments. EfficientNet introduces a method called Compound Scaling, which involves uniformly scaling the network depth, width, and image resolution to achieve improved performance. The key structures that contribute to its effectiveness are as follows:

1)	Uniform Scaling of Width, Depth, and Resolution: EfficientNet employs a unified scaling strategy that simultaneously scales the model's width, depth, and input image resolution. This strategy balances and optimizes the model's capacity, resulting in improved performance across different tasks while maintaining efficiency.

2)	Inverted Residuals: EfficientNet incorporates the inverted residual structure inspired by MobileNetV2. It utilizes a combination of lightweight depthwise convolutions and pointwise convolutions to achieve an efficient network design. The inverted residual structure reduces computational and parameter complexity while preserving model accuracy.

3)	MBConv (Mobile Inverted Residual Bottleneck) Blocks: EfficientNet utilizes MBConv blocks as its fundamental building units. MBConv blocks combine lightweight depthwise convolutions, pointwise convolutions, and shortcut connections, enhancing both model efficiency and feature representation capabilities. These blocks enable EfficientNet to extract powerful features and maintain good information flow.

4)	Attention Mechanism: EfficientNet introduces attention mechanisms in the later stages of the network to enhance the focus on important features. By adaptively adjusting the weights of feature maps, the attention mechanism improves the model's perception of crucial regions and features within the image, further enhancing classification performance.

\subsubsection{RegNet}
RegNet (Regularized Evolution of Architectures)~\citep{radosavovic2020designing} is a method for automatically searching neural network architectures proposed by the research team at Stanford University. RegNet aims to guide the search and development of neural networks by designing a set of rules to achieve efficient and high-performing architectures. Its exceptional performance in classification tasks can be attributed to key structures and characteristics derived from its unique design principles and adjustment strategies. The following highlights the structures and features that make RegNet excel in classification tasks:

1)	Rule-based growth of network depth and width: RegNet adopts a strategy of rule-based growth by increasing the depth and width of the network to enhance its expressive power. In contrast to traditional networks, RegNet ensures model balance and adjustability by increasing depth and width at the same growth rate across different levels.

2)	Scaling strategy for network depth and width: RegNet employs scaling coefficients to adjust the growth rate of network depth and width. This scaling strategy enables the network to flexibly adapt to different computational resources and task requirements, resulting in efficient network designs.

3)	Regularization mechanisms: RegNet incorporates regularization mechanisms such as Dropout and Weight Decay to reduce overfitting and improve model generalization. These regularization techniques facilitate better adaptation of the model to training data during the training process and reduce errors on test data.

4)	Network architecture search and optimization: RegNet automates the process of searching and selecting network architectures through network architecture search and optimization algorithms such as AutoML and Neural Architecture Search (NAS). This ensures that RegNet achieves high performance and efficiency in classification tasks.

Therefore, RegNet's outstanding performance in classification tasks can be attributed to its unique design principles and adjustment strategies. The rule-based growth of network depth and width, scaling strategy for depth and width, regularization mechanisms, and automated network architecture search and optimization contribute to its remarkable performance and efficiency.

\begin{figure*}[th]
\centering
\begin{tabular}{c}
\includegraphics[width=0.99\linewidth]{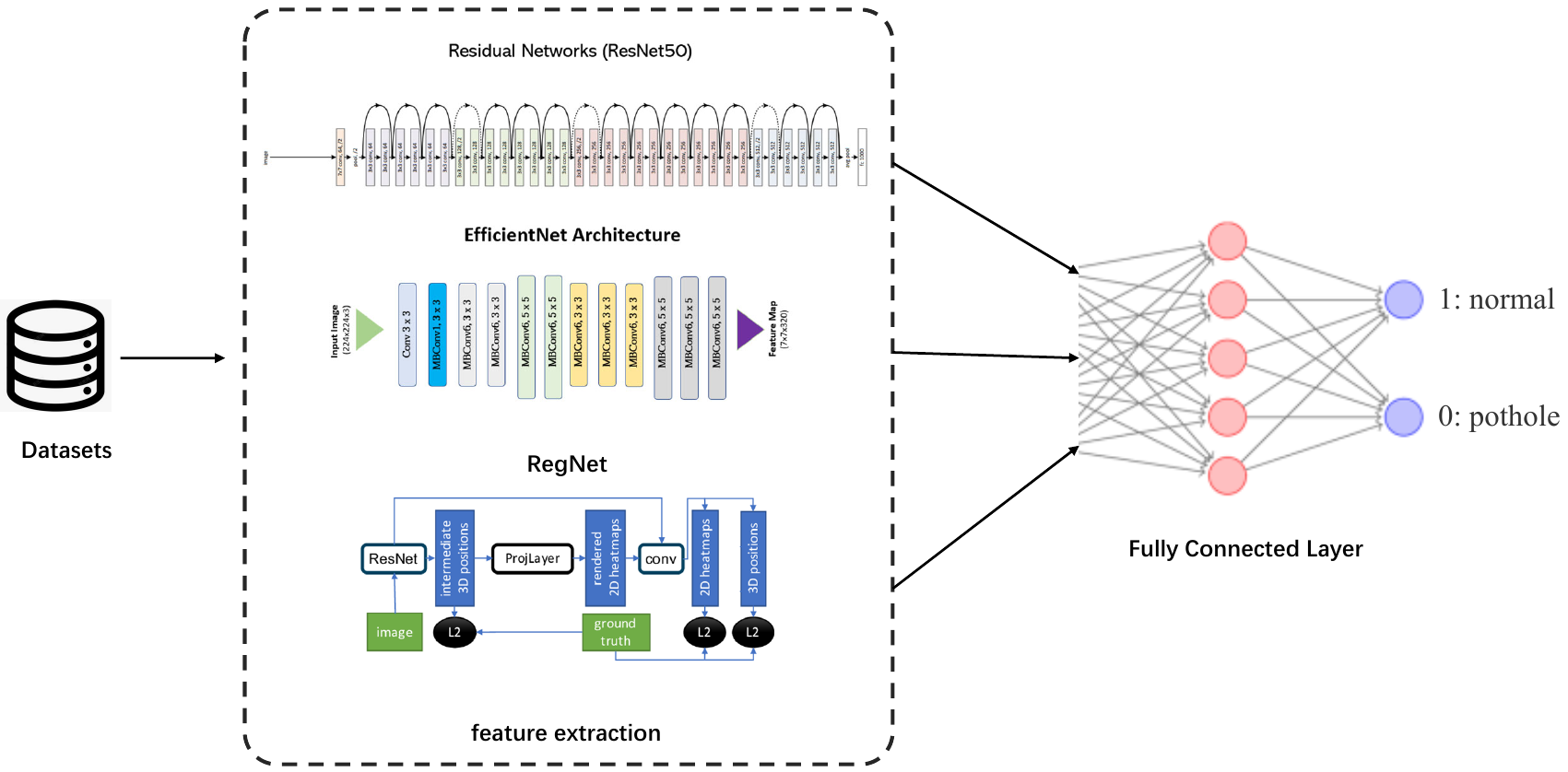}\\
% \vspace{}
\end{tabular}
\caption{\small{ResNet50-EfficientNet-RegNet model structure diagram.}}
\label{fig:3}
\end{figure*}

\subsubsection{Fully Connected Layer}
The units in convolutional neural networks (CNNs) that transform feature maps into one-dimensional vectors are referred to as fully connected layers. These layers encompass all the features of an image, as they establish connections between every node in the current layer and the preceding layer of the network. By extracting global features from the image, the fully connected layers enable subsequent classification. Consequently, it is common for fully connected layers to appear at the end of CNNs. After the image undergoes convolution and pooling operations to extract features, the resulting features are then mapped to the label space of the samples through the utilization of fully connected layers.

Therefore, in this study, we propose a road pothole detection and recognition model. The model architecture consists of an input layer, ResNet50 layer, EfficientNet layer, RegNet layer, and a fully connected output layer (5120 x 2). The overall structure of the road pothole detection model based on ResNet50-EfficientNet-RegNet is illustrated in \figurename~\ref{fig:3}.

\subsection{Feature Extraction and Visualization}
\begin{figure}[h]
\centering
\begin{tabular}{c}
\includegraphics[width=0.99\linewidth]{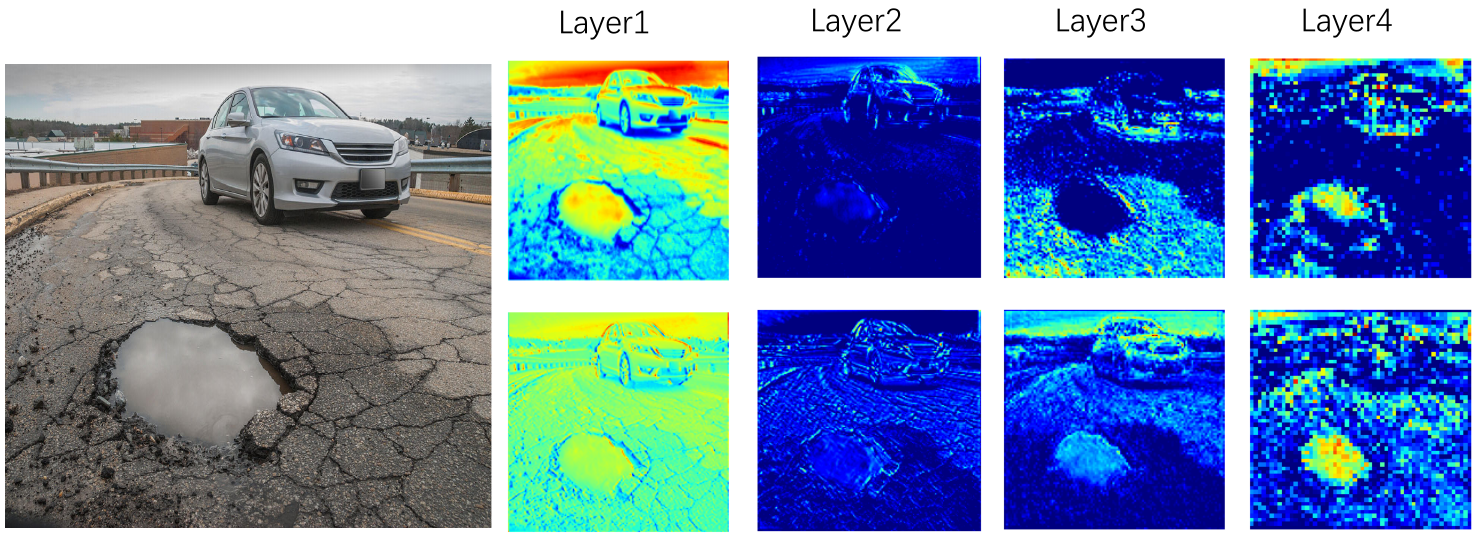}\\
% \vspace{}
\includegraphics[width=0.99\linewidth]{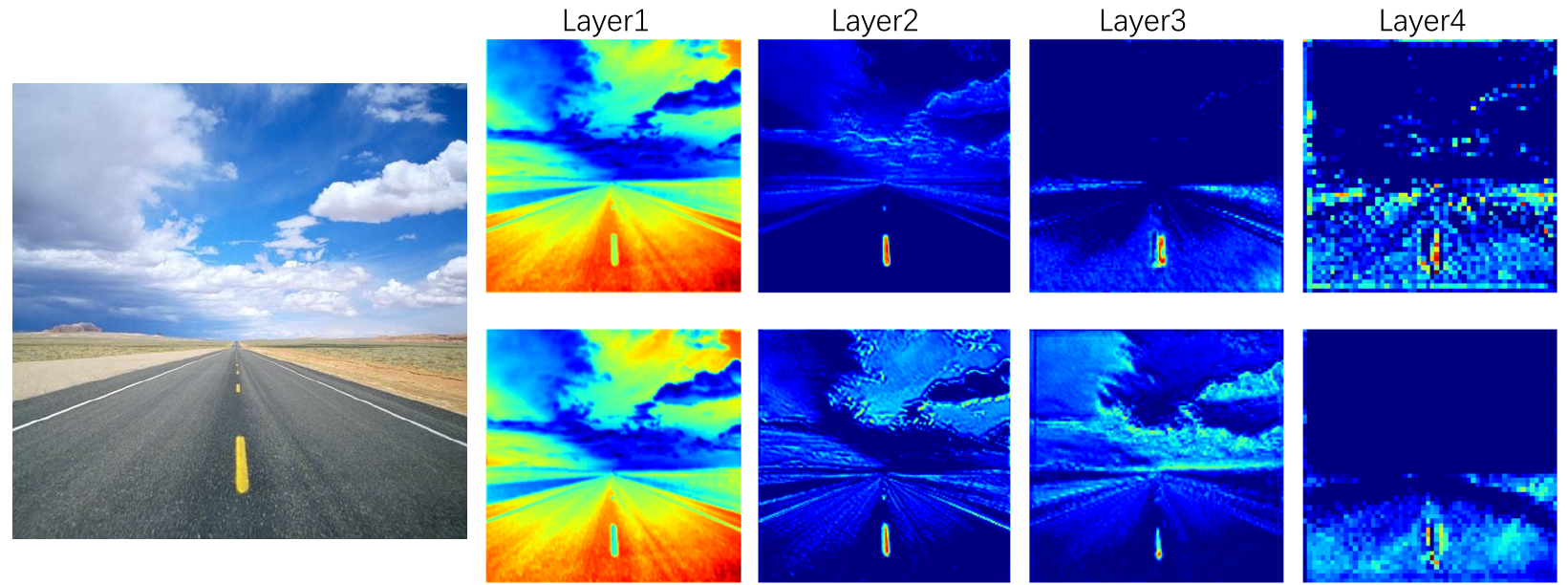}\\
\end{tabular}
\caption{\small{Network middle layer feature heat map}}
\label{fig:4}
\end{figure}

The classification performance of the classification network depends on the quality of the learned image features. Convolutional neural networks (CNNs) extract feature maps and feature vectors through convolution and pooling operations. In the case of road pothole data classification, our approach is based on ResNet50-EfficientNet-RegNet. By training and fine-tuning the network, we obtained optimal network weights for extracting relevant features of road potholes. In this study, the convolutional layers of ResNet50, EfficientNet, and RegNet are utilized to extract features from the input data. \figurename~\ref{fig:4} shows the feature heatmaps of the intermediate layers in the network. From \figurename~\ref{fig:4}, it can be observed that as the number of network layers increases, the features related to road potholes become more evident.

Both T-SNE (t-Distributed Stochastic Neighbor Embedding) and PCA (Principal Component Analysis) are dimensionality reduction techniques used for visualizing high-dimensional data or extracting key features from the data. The T-SNE and PCA feature visualization images in Figure 5 represent the dimensionality reduction of road pothole images. Each point represents an image, with the x-axis and y-axis representing the variance of the first principal component and the second principal component, respectively. The red color indicates features of normal road images, while the blue color represents features of road images with potholes.
\begin{figure*}[h]
\centering
\begin{subfigure}{0.5\linewidth}
    \centering
    \includegraphics[width=\linewidth]{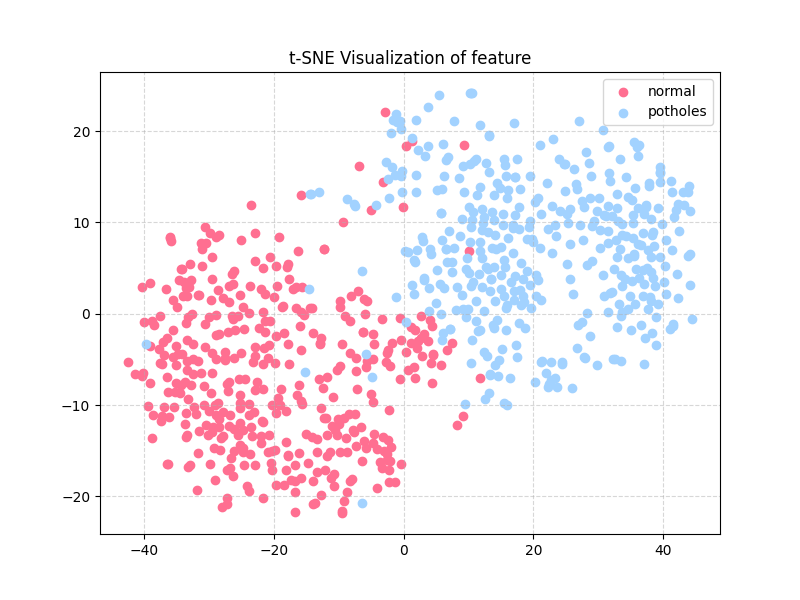}
    \caption{T-SNE}
\end{subfigure}%
\begin{subfigure}{0.5\linewidth}
    \centering
    \includegraphics[width=\linewidth]{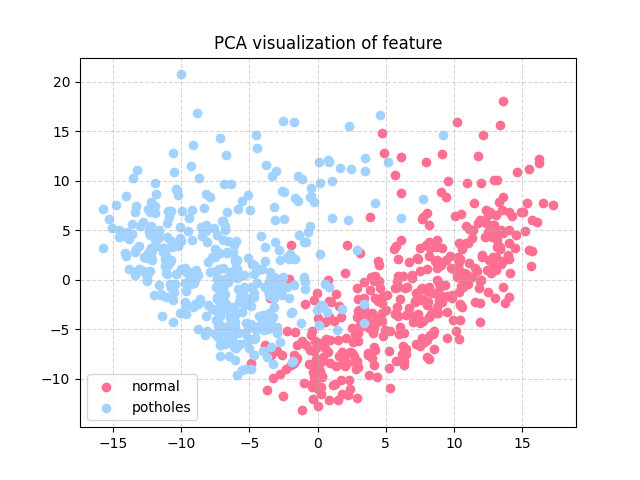}
    \caption{PCA}
\end{subfigure}
\caption{\small{T-SNE and PCA feature dimensionality reduction visualization diagram.}}
\label{fig:5}
\end{figure*}

From \figurename~\ref{fig:5}, it is evident that after the feature extraction process by the network, the features related to road potholes and non-pothole features have been effectively separated. This clear distinction between the two types of features provides a solid foundation for subsequent image classification tasks.

\begin{figure}[th]
\centering
\begin{tabular}{c}
\includegraphics[width=0.99\linewidth]{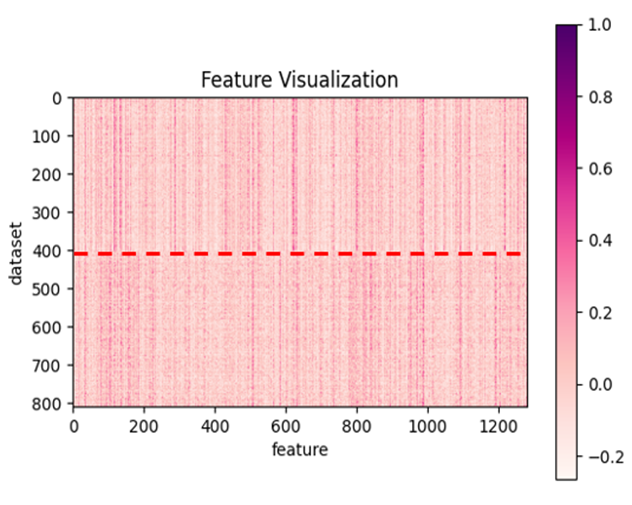}\\
% \vspace{}
\end{tabular}
\caption{\small{Visualization diagram of extracted features for model.}}
\label{fig:6}
\end{figure}

\figurename~\ref{fig:6} depicts the feature visualization of the EfficientNet model. The y-axis represents the 810 images from the training set, where the range 0-405 corresponds to normal road images and the range 405-800 represents road images with potholes. The x-axis represents the number of features extracted by the model, with EfficientNet, ResNet, and RegNet extracting 1280, 2048, and 2016-dimensional features, respectively. The scale on the right side of the graph, ranging from -0.2 to 1.0, represents the magnitude of the feature values.

From the graph, it is evident that the differences within normal road images and within road images with potholes are relatively small for the same feature. However, there is a noticeable distinction between normal road and road with pothole images. This indicates that the model is capable of classifying normal road images and road images with potholes effectively.

\section{Data processing}
To meet the requirements of road pothole detection, we collected a dataset of road potholes through manual collection and internet sources. This dataset encompasses diverse content, including different time periods, lighting conditions, road segments, and varying numbers of potholes. To satisfy the data training requirements of deep learning, we utilized the LabelImg software to annotate the dataset specifically for pothole detection. The corresponding labels were assigned as follows: road images without potholes were labeled as 1, while road images with potholes were labeled as 0. In the end, we compiled an original dataset consisting of 450 images of normal road surfaces and 450 images of road surfaces with potholes. Furthermore, we performed standardization, normalization, and data augmentation techniques on the original image data to enhance the training process.

\subsection{Standardization and Normalization}
In this study, the min-max normalization method, also known as the range normalization, was employed. It involves a linear transformation of the original data to map the resulting values within the range of [0,1]. The transformation function is defined as follows:
\begin{equation}
\begin{gathered}
x'=\frac{x-x_{min}}{x_{max}-x_{min}},
\end{gathered}
\end{equation}
In the equation, $x$ represents the original pixel value, $x'$ represents the normalized pixel value, $x_max$ represents the maximum pixel value in the sample data, and $x_min$ is set to 0 in this case since the pixel values are all positive. The normalization transformation function used is as follows:
\begin{equation}
\begin{gathered}
x'=\frac{x}{x_{max}}.
\end{gathered}
\end{equation}

\subsection{Image augmentation}
Since the size of road images in the dataset is uncontrollable and uniform size has a certain impact on the effectiveness of the classification network, it is important to ensure consistency in the size of the dataset images. Therefore, we employed the resize method to standardize the size of the road images in the multi-scene dataset to 224x224 pixels.

\begin{figure}[h]
\centering
\begin{tabular}{c}
\includegraphics[width=0.99\linewidth]{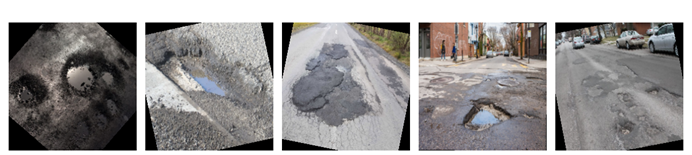}\\
% \vspace{}
\includegraphics[width=0.99\linewidth]{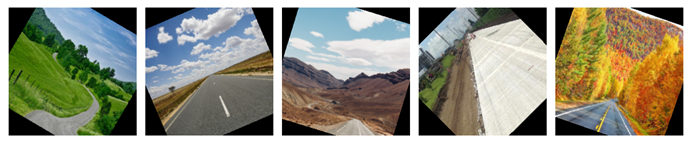}\\
\end{tabular}
\caption{\small{Image display after random rotation}}
\label{fig:1}
\end{figure}

In order to enhance the model's generalization ability, this study employed two data augmentation techniques: random rotation and random horizontal flipping. By introducing these techniques, the training dataset was augmented with diverse variations, which helped to mitigate the risk of overfitting to some extent. Specifically, random rotation was applied with angles ranging from -45° to 45°. This introduced a range of rotation variations to the training dataset, further diversifying the data and reducing the likelihood of overfitting. \figurename~\ref{fig:1} shows random rotation of the images.

\subsection{Partition of the Dataset}
After processing the dataset, a total of 900 images were obtained. The data was divided into training and testing sets in a ratio of 9:1, resulting in 810 images for the training set and 90 images for the testing set. The partition of the dataset is presented in \tablename~\ref{tab:1}.
\begin{table}
\centering
\caption{\small{Partition of the Dataset}}
\begin{tabular}{l c c}
\toprule
  & normal road & pothole road \\
\midrule
testing set & 45 & 45 \\
training set & 405 & 405 \\
\bottomrule
\end{tabular}
\label{tab:1}
\end{table}

\section{Experiments}\label{sec4}
\subsection{Experimental Setup}
In this study, the ResNet50-EfficientNet-RegNet model was built using PyTorch. During the model construction process, ReLU activation function was employed for both ResNet50 and RegNet, while EfficientNet utilized the Swish activation function. The Swish activation function combines a linear part similar to ReLU and a non-linear part similar to Sigmoid. Its advantage lies in introducing richer non-linearity while maintaining linear characteristics, thus enhancing the model's expressive power compared to ReLU.

The model's loss function has been modified by introducing the L1 loss function in addition to the original Binary Crossentropy loss function. This enhancement aims to improve the model's robustness, generalization ability, feature selection, and sparsity, consequently optimizing its performance. The Binary Crossentropy loss function primarily measures the discrepancy between predicted results and true labels, while the L1 loss function promotes the generation of sparser solutions by the model. Therefore, the incorporation of the L1 loss function enhances the model's robustness to outliers or noisy data, reducing its sensitivity towards such instances.

 The optimizer employed in the model is the popular Adam optimization algorithm. Adam algorithm is known to update network weights more efficiently compared to stochastic gradient descent (SGD) algorithm, making it suitable for training models with large datasets and a high number of parameters.
 
For regularization, L2 regularization, also known as weight decay, was applied. L2 regularization reduces model complexity by limiting the freedom of weights, resulting in smoother models that exhibit better tolerance to small perturbations in the training data. Additionally, L2 regularization facilitates better generalization to unseen data.

\begin{table}[h]
\centering
\caption{\small{Model hyperparameters}}
\begin{tabular}{l c}
\toprule
Model hyperparameters & Parameter value \\
\midrule
Batch\_size & 30 \\
hidden\_units & 5344 \\
num\_labels & 2 \\
Epochs & 5 \\
Dropout\_rate & 0.5 \\
learning\_rate & 0.01 \\
learning\_rate\_decay & 1 \\
lr\_decat\_step & 400 \\
weight\_decay & 5e-4 \\
momentum & 0.9 \\
\bottomrule
\end{tabular}
\label{tab:2}
\end{table}

The hyperparameters of the model are shown in \tablename~\ref{tab:2}. Specifically, Batch\_size represents the number of samples processed in each training iteration.; hidden\_units denotes the number of hidden units in the fully connected layer.; num\_labels indicates the number of output categories in the model. Epochs corresponds to the total number of training iterations. Dropout\_rate signifies the rate at which random neurons are dropped during each training step; learning\_rate represents the initial learning rate of the model; learning\_rate\_decay denotes the coefficient for decaying the learning rate. lr\_decat\_step signifies the number of iterations after which the learning rate decays. weight\_decay represents the regularization coefficient, while momentum denotes the coefficient for the momentum optimization method.

The specific training steps for the road pothole detection model based on ResNet50-EfficientNet-RegNet are as follows:

1)	The input layer receives the corresponding feature parameters' data as input.

2)	The convolutional layers in ResNet50, EfficientNet, and RegNet extract features from the input data, selecting important feature information. Specifically, ResNet50 extracts features with a dimensionality of 2048, EfficientNet extracts features with a dimensionality of 1280, and RegNet extracts features with a dimensionality of 2016.

3)	The fully connected layer (5120*2) processes the output data from ResNet50, EfficientNet, and RegNet, enabling the classification of road pothole images.

4)	The model's loss function is calculated to determine if the training termination condition has been met. If it has, the model is saved; otherwise, the model parameters are updated for the next training iteration.

\begin{figure*}[th]
\centering
\begin{subfigure}{0.24\linewidth}
    \centering
    \includegraphics[width=\linewidth]{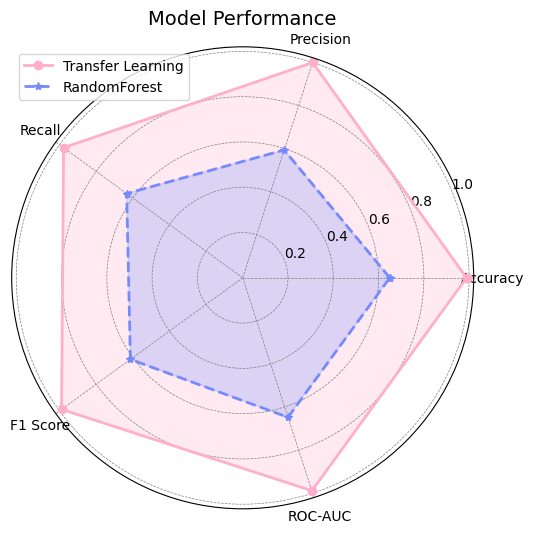}
    \caption{Random Forest}
\end{subfigure}%
\begin{subfigure}{0.24\linewidth}
    \centering
    \includegraphics[width=\linewidth]{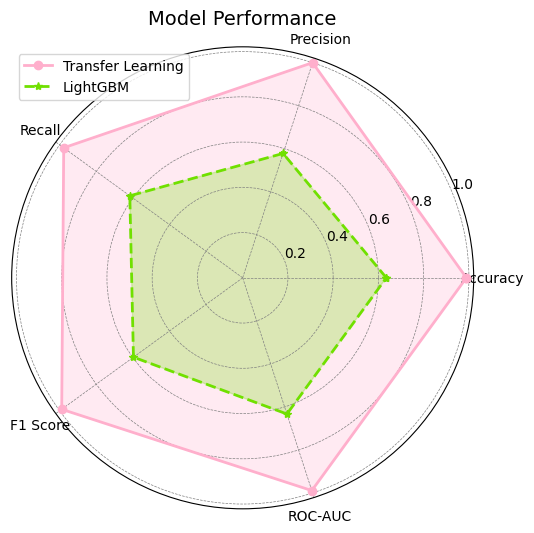}
    \caption{LightGBM}
\end{subfigure}
\begin{subfigure}{0.24\linewidth}
    \centering
    \includegraphics[width=\linewidth]{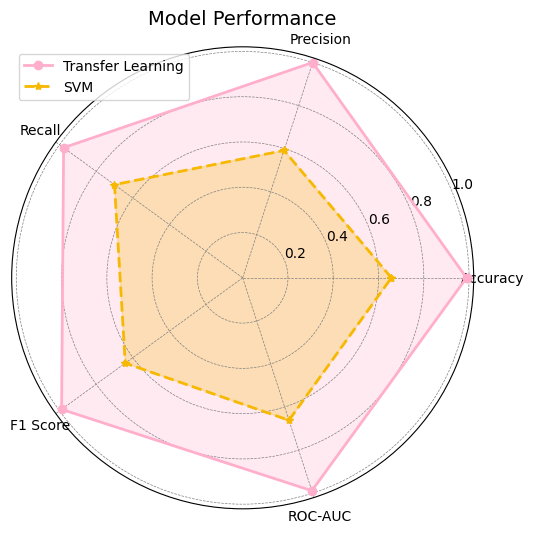}
    \caption{SVM}
\end{subfigure}
\begin{subfigure}{0.24\linewidth}
    \centering
    \includegraphics[width=\linewidth]{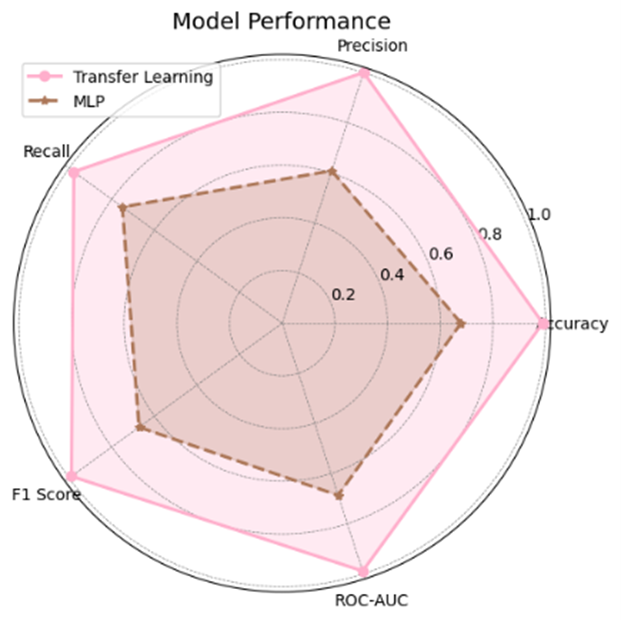}
    \caption{MLP}
\end{subfigure}
\caption{\small{Radar chart of classification effects comparing each model with transfer learning.}}
\label{fig:7}
\end{figure*}

\subsection{Result and Analysis}
Based on the established and trained model, we evaluate its performance using both metric evaluation and comparative evaluation methods. The metric evaluation includes commonly used accuracy assessment metrics in the field of object detection, such as Accuracy, Recall, Precision, F1-score, and the model speed evaluation metric FPS (Frames Per Second). The following are the calculation formulas for some indicators. 
\begin{equation}
\begin{gathered}
Accuracy=\frac{Right simples}{All simples},
\end{gathered}
\end{equation}

\begin{equation}
\begin{gathered}
Recall = \frac{True Positives}{True Positives+False Negatives},
\end{gathered}
\end{equation}

\begin{equation}
\begin{gathered}
Precision = \frac{True Positives}{True Positives+False Positives},
\end{gathered}
\end{equation}

\begin{equation}
\begin{gathered}
F1-score=\frac{2\cdot Precision\cdot Recall}{Precision+Recall},
\end{gathered}
\end{equation}

The comparative evaluation involves comparing the performance of the model with other algorithms, namely Random Forest~\citep{rigatti2017random}, MLP~\citep{taud2018multilayer}, SVM~\citep{noble2006support}, LightGBM~\citep{ke2017lightgbm}, and the ResNet method in deep learning. We conduct a comparative analysis by applying these models and transfer learning models to the given dataset for pothole road image recognition. The training and testing datasets are divided in a 9:1 ratio. The classification results of each algorithm are visualized in the following figures.

\begin{figure*}[th]
\centering
\begin{subfigure}{0.5\linewidth}
    \centering
    \includegraphics[width=\linewidth]{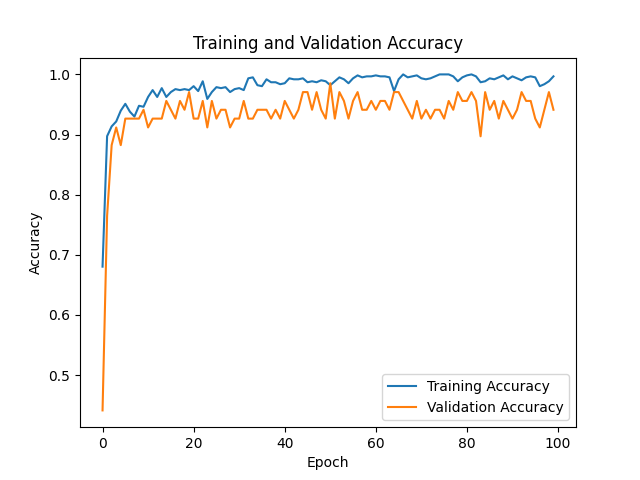}
    \caption{Accuracy change chart}
\end{subfigure}%
\begin{subfigure}{0.5\linewidth}
    \centering
    \includegraphics[width=\linewidth]{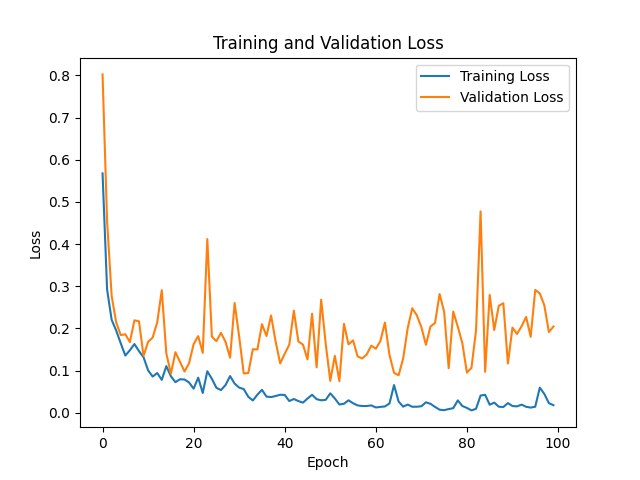}
    \caption{Loss value change chart}
\end{subfigure}
\caption{\small{ResNet accuracy and loss value change chart.}}
\label{fig:8}
\end{figure*}

\figurename~\ref{fig:7} displays a radar chart illustrating the performance of various models on the dataset. The chart compares our proposed ResNet50-EfficientNet-RegNet model with traditional models such as Random Forest, LightGBM, SVM, and MLP, based on five evaluation metrics: Accuracy, Precision, Recall, F1 Score, and the area under the ROC curve (ROC-AUC).

The results clearly demonstrate that the ResNet50-EfficientNet-RegNet model outperforms the traditional models in all aspects. Specifically, the transfer learning model achieves an accuracy of 0.99, which is close to 1. In contrast, the Random Forest, LightGBM, SVM, and MLP models achieve lower accuracy values of 0.65, 0.63, 0.66, and 0.68, respectively. These values only show a slight improvement over random classification (0.5). Therefore, it is evident that our proposed ResNet50-EfficientNet-RegNet model significantly excels in image classification compared to traditional machine learning models.

\figurename~\ref{fig:8} presents the variation of accuracy and loss values on the training and testing sets during the training process of the deep learning ResNet network. From the figure, it can be observed that the accuracy of the ResNet network gradually improves on both the training and testing sets. However, after approximately 10 epochs of training, the accuracy values start to fluctuate and fail to converge effectively. Consequently, it indicates that the model reaches a convergence point around the 10th epoch, and the training performance is not ideal.

Moreover, during the training process, the loss functions of both the training and testing sets exhibit a downward trend. However, while the training loss steadily decreases, the testing loss begins to fluctuate and fails to reach the same level as the training loss. This phenomenon may be attributed to the complexity of the model or a limited amount of training data, resulting in overfitting of the model to the training set or insufficient model capacity to generalize well to the validation set.
\begin{figure*}[h]
\centering
\begin{tabular}{c}
\includegraphics[width=0.99\linewidth]{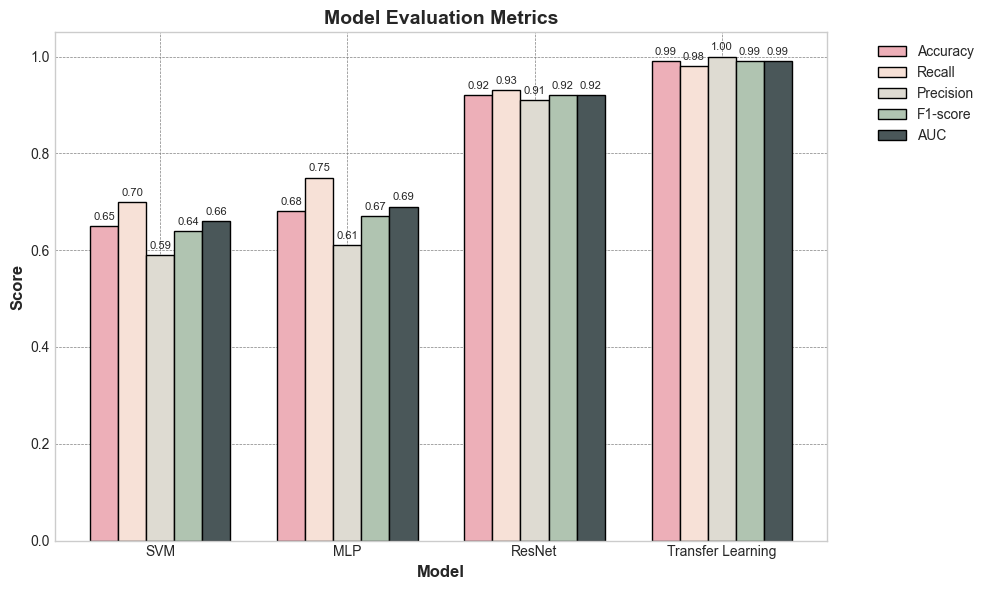}\\
% \vspace{}
\end{tabular}
\caption{\small{Model evaluation metrics histogram.}}
\label{fig:9}
\end{figure*}

\figurename~\ref{fig:9} represents a histogram depicting the evaluation metrics of various models in the context of road image pothole detection. The x-axis represents four models: SVM, MLP, ResNet, and Transfer Learning, while the y-axis represents the numerical values of evaluation metrics such as Accuracy, Recall, Precision, and others. From the histogram, it is evident that both ResNet and Transfer Learning models exhibit significant improvements in all evaluation metrics compared to the traditional SVM and MLP models. Particularly, our Transfer Learning model outperforms the ResNet model in image classification. It achieves an Accuracy of 0.99, Recall of 0.98, F1 score of 0.99, and AUC of 0.99, all of which are close to 1. Additionally, it attains a Precision of 1.
\begin{figure*}[h]
\centering
\begin{tabular}{c}
\includegraphics[width=0.99\linewidth]{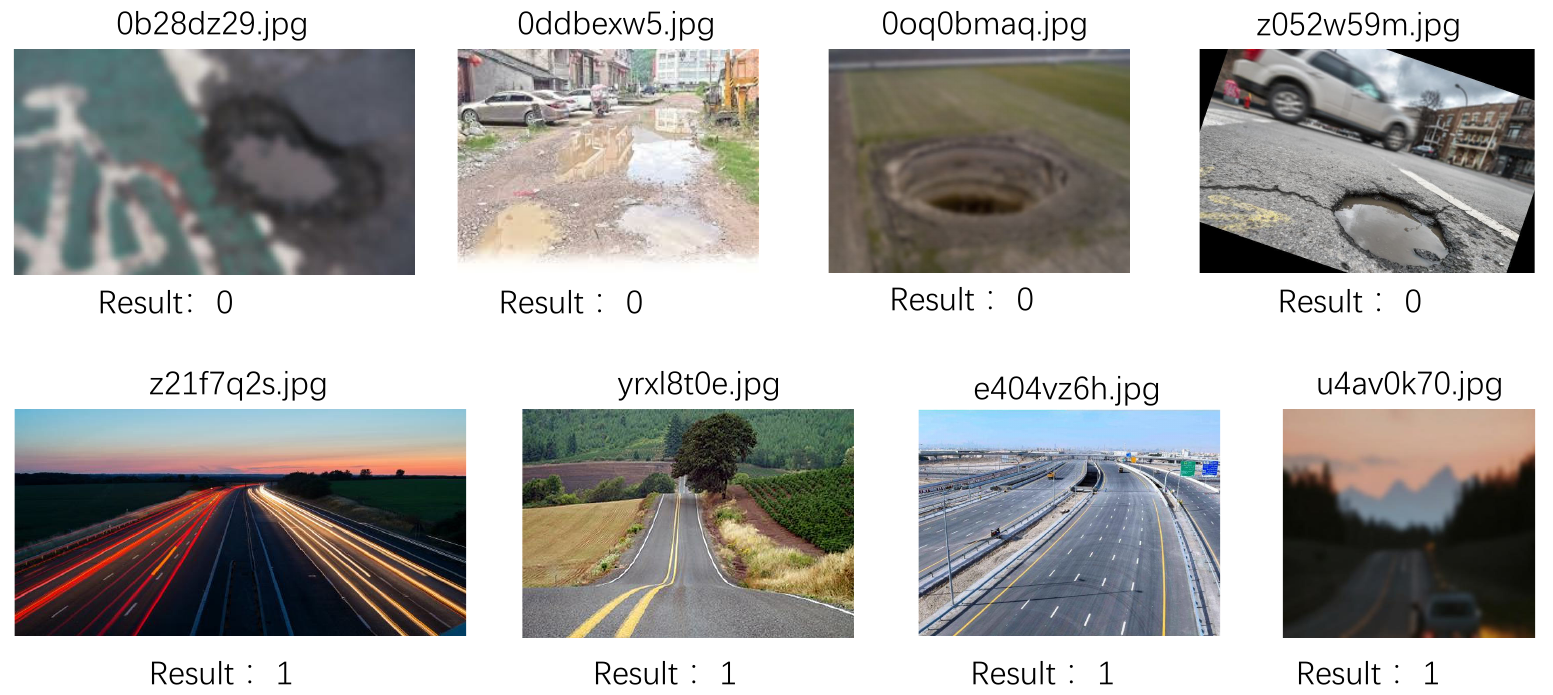}\\
% \vspace{}
\end{tabular}
\caption{\small{Partial test set recognition results.}}
\label{fig:10}
\end{figure*}

\begin{table*}
\centering
\caption{\small{Model efficiency comparison}}
\begin{tabular}{l c c c c}
\toprule
Model & ResNet & EfficientNet & GoogLeNet & Transfer Learning \\
\midrule
FPS & 0.02 & 0.03 & 0.08 & 1.03 \\
\bottomrule
\end{tabular}
\label{tab:3}
\end{table*}
\tablename~\ref{tab:3} showcases the FPS values of four models, namely ResNet, EfficientNet, GoogLeNet~\citep{szegedy2015going}, and Transfer Learning, using the same test dataset and measured in milliseconds on a single GPU. From the table, it is evident that our Transfer Learning model achieves an FPS of 1.03, which is significantly higher compared to the ResNet, EfficientNet, and GoogLeNet models with FPS values of 0.02, 0.03, and 0.08, respectively. These results indicate that our proposed Transfer Learning model exhibits a faster image recognition speed.

Upon comparing the aforementioned models, we discover that the predictive performance of deep learning models is more accurate than that of machine learning models. This highlights the advantages of deep learning models. Our deep learning model effectively leverages the strengths of ResNet50, EfficientNet, and RegNet networks individually, and integrates them together to further enhance the accuracy of classification. However, the advantage of machine learning lies in its ability to train models without requiring an extensive amount of data, while still ensuring a certain level of accuracy.

\figurename~\ref{fig:10} presents a subset of road images selected from the test set, which exhibit blurriness, non-standard elements, and misorientation. These types of images pose certain difficulties in the task of pothole detection on roads. However, it is evident from the figure that our model achieves a high level of accuracy when handling such images.

In order to better evaluate the generalization ability of our model, we extensively collected high-quality datasets from sources such as Google, GitHub, and other websites. These datasets were combined with the previously mentioned dataset, resulting in a large-scale test set comprising 900 images. This test set consists of 270 images depicting normal road conditions and 630 images featuring road conditions with potholes. After conducting feature extraction, model training, and classification, we achieved a classification accuracy of 98.89\% (890/900) on this test set. This remarkable performance once again confirms the generalization ability and accuracy of our model.

\section{Conclusion}
This paper proposes a transfer learning-based algorithm to address the challenges faced by existing object detection algorithms in accurately detecting and slow detection speed of road potholes. The algorithm effectively leverages the individual strengths of ResNet50, EfficientNet, and RegNet networks and integrates them together. Through the utilization of custom fully connected layers and improvements in the loss function, the algorithm achieves further enhancement in classification accuracy. Experimental results demonstrate that the pro-posed transfer learning-based ResNet50-EfficientNet-RegNet model exhibits high performance across various dimensions in the task of road pothole detection, surpassing the performance of other models. This indicates a significant advantage of the proposed model in road pothole detection tasks. Moving forward, further research will be conducted to explore how to maintain detection speed while improving detection accuracy in the face of multiple-scene road pothole environments, aiming to achieve better detection outcomes.

% \section*{Declarations}

% \textbf{Competing Interests.} 
% The authors declared that they have no competing interests.

% \vspace{1\baselineskip}
% \noindent
% \textbf{Authors' contributions.} 
% XQ developed the overall research goals and objectives and subsequently wrote the main manuscript text. HM took on the responsibility of analyzing and organizing the data, as well as preparing the figures. All authors reviewed the manuscript.

% \vspace{1\baselineskip}
% \noindent
% \textbf{Funding.} 
% This work was supported by Open Research Project of The Hubei Key Laboratory of Intelligent Geo-Information Processing(KLIGIP-2022-B13).

% \vspace{1\baselineskip}
% \noindent
% \textbf{Availability of data and materials.} 
% Relevant data for this work are available to the extent reasonably requested by contacting the corresponding author.

% \bibliographystyle{sn-apa}
\bibliography{main}% common bib file

\begin{thebibliography}{}
\renewcommand{\doi}[1]{\url{https://doi.org/#1}}
\bibcommenthead

\bibitem [\protect \citeauthoryear {%
Fan%
, Ozgunalp%
, Hosking%
, Liu%
\BCBL {}\ \BBA {} Pitas%
}{%
Fan%
\ \protect \BOthers {.}}{%
{\protect \APACyear {2019}}%
}]{%
fan2019pothole}
\APACinsertmetastar {%
fan2019pothole}%
\begin{APACrefauthors}%
Fan, R.%
, Ozgunalp, U.%
, Hosking, B.%
, Liu, M.%
\BCBL {} Pitas, I.%
\end{APACrefauthors}%
\unskip\
\newblock
\APACrefYearMonthDay{2019}{}{}.
\newblock
{\BBOQ}\APACrefatitle {Pothole detection based on disparity transformation and road surface modeling} {Pothole detection based on disparity transformation and road surface modeling}.{\BBCQ}
\newblock
\APACjournalVolNumPages{IEEE Transactions on Image Processing}{29}{}{897--908,}
\newblock

\newblock

\PrintBackRefs{\CurrentBib}

\bibitem [\protect \citeauthoryear {%
Fan%
, Ozgunalp%
, Wang%
, Liu%
\BCBL {}\ \BBA {} Pitas%
}{%
Fan%
\ \protect \BOthers {.}}{%
{\protect \APACyear {2021}}%
}]{%
fan2021rethinking}
\APACinsertmetastar {%
fan2021rethinking}%
\begin{APACrefauthors}%
Fan, R.%
, Ozgunalp, U.%
, Wang, Y.%
, Liu, M.%
\BCBL {} Pitas, I.%
\end{APACrefauthors}%
\unskip\
\newblock
\APACrefYearMonthDay{2021}{}{}.
\newblock
{\BBOQ}\APACrefatitle {Rethinking road surface 3-d reconstruction and pothole detection: From perspective transformation to disparity map segmentation} {Rethinking road surface 3-d reconstruction and pothole detection: From perspective transformation to disparity map segmentation}.{\BBCQ}
\newblock
\APACjournalVolNumPages{IEEE Transactions on Cybernetics}{52}{7}{5799--5808,}
\newblock

\newblock

\PrintBackRefs{\CurrentBib}

\bibitem [\protect \citeauthoryear {%
Girshick%
}{%
Girshick%
}{%
{\protect \APACyear {2015}}%
}]{%
girshick2015fast}
\APACinsertmetastar {%
girshick2015fast}%
\begin{APACrefauthors}%
Girshick, R.%
\end{APACrefauthors}%
\unskip\
\newblock
\APACrefYearMonthDay{2015}{}{}.
\newblock
{\BBOQ}\APACrefatitle {Fast r-cnn} {Fast r-cnn}.{\BBCQ}
\newblock
 \APACrefbtitle {Proceedings of the IEEE international conference on computer vision} {Proceedings of the ieee international conference on computer vision}\ (\BPGS\ 1440--1448).
\PrintBackRefs{\CurrentBib}

\bibitem [\protect \citeauthoryear {%
Girshick%
, Donahue%
, Darrell%
\BCBL {}\ \BBA {} Malik%
}{%
Girshick%
\ \protect \BOthers {.}}{%
{\protect \APACyear {2014}}%
}]{%
girshick2014rich}
\APACinsertmetastar {%
girshick2014rich}%
\begin{APACrefauthors}%
Girshick, R.%
, Donahue, J.%
, Darrell, T.%
\BCBL {} Malik, J.%
\end{APACrefauthors}%
\unskip\
\newblock
\APACrefYearMonthDay{2014}{}{}.
\newblock
{\BBOQ}\APACrefatitle {Rich feature hierarchies for accurate object detection and semantic segmentation} {Rich feature hierarchies for accurate object detection and semantic segmentation}.{\BBCQ}
\newblock
 \APACrefbtitle {Proceedings of the IEEE conference on computer vision and pattern recognition} {Proceedings of the ieee conference on computer vision and pattern recognition}\ (\BPGS\ 580--587).
\PrintBackRefs{\CurrentBib}

\bibitem [\protect \citeauthoryear {%
He%
, Gkioxari%
, Doll{\'a}r%
\BCBL {}\ \BBA {} Girshick%
}{%
He%
\ \protect \BOthers {.}}{%
{\protect \APACyear {2017}}%
}]{%
he2017mask}
\APACinsertmetastar {%
he2017mask}%
\begin{APACrefauthors}%
He, K.%
, Gkioxari, G.%
, Doll{\'a}r, P.%
\BCBL {} Girshick, R.%
\end{APACrefauthors}%
\unskip\
\newblock
\APACrefYearMonthDay{2017}{}{}.
\newblock
{\BBOQ}\APACrefatitle {Mask r-cnn} {Mask r-cnn}.{\BBCQ}
\newblock
 \APACrefbtitle {Proceedings of the IEEE international conference on computer vision} {Proceedings of the ieee international conference on computer vision}\ (\BPGS\ 2961--2969).
\PrintBackRefs{\CurrentBib}

\bibitem [\protect \citeauthoryear {%
He%
, Zhang%
, Ren%
\BCBL {}\ \BBA {} Sun%
}{%
He%
\ \protect \BOthers {.}}{%
{\protect \APACyear {2016}}%
}]{%
he2016deep}
\APACinsertmetastar {%
he2016deep}%
\begin{APACrefauthors}%
He, K.%
, Zhang, X.%
, Ren, S.%
\BCBL {} Sun, J.%
\end{APACrefauthors}%
\unskip\
\newblock
\APACrefYearMonthDay{2016}{}{}.
\newblock
{\BBOQ}\APACrefatitle {Deep residual learning for image recognition} {Deep residual learning for image recognition}.{\BBCQ}
\newblock
 \APACrefbtitle {Proceedings of the IEEE conference on computer vision and pattern recognition} {Proceedings of the ieee conference on computer vision and pattern recognition}\ (\BPGS\ 770--778).
\PrintBackRefs{\CurrentBib}

\bibitem [\protect \citeauthoryear {%
Ke%
\ \protect \BOthers {.}}{%
Ke%
\ \protect \BOthers {.}}{%
{\protect \APACyear {2017}}%
}]{%
ke2017lightgbm}
\APACinsertmetastar {%
ke2017lightgbm}%
\begin{APACrefauthors}%
Ke, G.%
, Meng, Q.%
, Finley, T.%
, Wang, T.%
, Chen, W.%
, Ma, W.%
\BDBL {}Liu, T\BHBI Y.%
\end{APACrefauthors}%
\unskip\
\newblock
\APACrefYearMonthDay{2017}{}{}.
\newblock
{\BBOQ}\APACrefatitle {Lightgbm: A highly efficient gradient boosting decision tree} {Lightgbm: A highly efficient gradient boosting decision tree}.{\BBCQ}
\newblock
\APACjournalVolNumPages{Advances in neural information processing systems}{30}{}{,}
\newblock

\newblock

\PrintBackRefs{\CurrentBib}

\bibitem [\protect \citeauthoryear {%
Kumar%
, Kalita%
, Singh%
\BCBL {}\ \protect \BOthers {.}}{%
Kumar%
\ \protect \BOthers {.}}{%
{\protect \APACyear {2020}}%
}]{%
kumar2020modern}
\APACinsertmetastar {%
kumar2020modern}%
\begin{APACrefauthors}%
Kumar, A.%
, Kalita, D.J.%
, Singh, V.P.%
\BCBL {}\ \BOthersPeriod {.}\end{APACrefauthors}%
\unskip\
\newblock
\APACrefYearMonthDay{2020}{}{}.
\newblock
{\BBOQ}\APACrefatitle {A modern pothole detection technique using deep learning} {A modern pothole detection technique using deep learning}.{\BBCQ}
\newblock
 \APACrefbtitle {2nd International Conference on Data, Engineering and Applications (IDEA)} {2nd international conference on data, engineering and applications (idea)}\ (\BPGS\ 1--5).
\PrintBackRefs{\CurrentBib}

\bibitem [\protect \citeauthoryear {%
Noble%
}{%
Noble%
}{%
{\protect \APACyear {2006}}%
}]{%
noble2006support}
\APACinsertmetastar {%
noble2006support}%
\begin{APACrefauthors}%
Noble, W.S.%
\end{APACrefauthors}%
\unskip\
\newblock
\APACrefYearMonthDay{2006}{}{}.
\newblock
{\BBOQ}\APACrefatitle {What is a support vector machine?} {What is a support vector machine?}{\BBCQ}
\newblock
\APACjournalVolNumPages{Nature biotechnology}{24}{12}{1565--1567,}
\newblock

\newblock

\PrintBackRefs{\CurrentBib}

\bibitem [\protect \citeauthoryear {%
Ouma%
\ \BBA {} Hahn%
}{%
Ouma%
\ \BBA {} Hahn%
}{%
{\protect \APACyear {2017}}%
}]{%
ouma2017pothole}
\APACinsertmetastar {%
ouma2017pothole}%
\begin{APACrefauthors}%
Ouma, Y.O.%
\BCBT {}\ \BBA {} Hahn, M.%
\end{APACrefauthors}%
\unskip\
\newblock
\APACrefYearMonthDay{2017}{}{}.
\newblock
{\BBOQ}\APACrefatitle {Pothole detection on asphalt pavements from 2D-colour pothole images using fuzzy c-means clustering and morphological reconstruction} {Pothole detection on asphalt pavements from 2d-colour pothole images using fuzzy c-means clustering and morphological reconstruction}.{\BBCQ}
\newblock
\APACjournalVolNumPages{Automation in Construction}{83}{}{196--211,}
\newblock

\newblock

\PrintBackRefs{\CurrentBib}

\bibitem [\protect \citeauthoryear {%
Pan%
\ \BBA {} Yang%
}{%
Pan%
\ \BBA {} Yang%
}{%
{\protect \APACyear {2009}}%
}]{%
pan2009survey}
\APACinsertmetastar {%
pan2009survey}%
\begin{APACrefauthors}%
Pan, S.J.%
\BCBT {}\ \BBA {} Yang, Q.%
\end{APACrefauthors}%
\unskip\
\newblock
\APACrefYearMonthDay{2009}{}{}.
\newblock
{\BBOQ}\APACrefatitle {A survey on transfer learning} {A survey on transfer learning}.{\BBCQ}
\newblock
\APACjournalVolNumPages{IEEE Transactions on knowledge and data engineering}{22}{10}{1345--1359,}
\newblock

\newblock

\PrintBackRefs{\CurrentBib}

\bibitem [\protect \citeauthoryear {%
Radosavovic%
, Kosaraju%
, Girshick%
, He%
\BCBL {}\ \BBA {} Doll{\'a}r%
}{%
Radosavovic%
\ \protect \BOthers {.}}{%
{\protect \APACyear {2020}}%
}]{%
radosavovic2020designing}
\APACinsertmetastar {%
radosavovic2020designing}%
\begin{APACrefauthors}%
Radosavovic, I.%
, Kosaraju, R.P.%
, Girshick, R.%
, He, K.%
\BCBL {} Doll{\'a}r, P.%
\end{APACrefauthors}%
\unskip\
\newblock
\APACrefYearMonthDay{2020}{}{}.
\newblock
{\BBOQ}\APACrefatitle {Designing network design spaces} {Designing network design spaces}.{\BBCQ}
\newblock
 \APACrefbtitle {Proceedings of the IEEE/CVF conference on computer vision and pattern recognition} {Proceedings of the ieee/cvf conference on computer vision and pattern recognition}\ (\BPGS\ 10428--10436).
\PrintBackRefs{\CurrentBib}

\bibitem [\protect \citeauthoryear {%
Ravi%
, Bullock%
\BCBL {}\ \BBA {} Habib%
}{%
Ravi%
\ \protect \BOthers {.}}{%
{\protect \APACyear {2020}}%
}]{%
ravi2020highway}
\APACinsertmetastar {%
ravi2020highway}%
\begin{APACrefauthors}%
Ravi, R.%
, Bullock, D.%
\BCBL {} Habib, A.%
\end{APACrefauthors}%
\unskip\
\newblock
\APACrefYearMonthDay{2020}{}{}.
\newblock
{\BBOQ}\APACrefatitle {Highway and airport runway pavement inspection using mobile lidar} {Highway and airport runway pavement inspection using mobile lidar}.{\BBCQ}
\newblock
\APACjournalVolNumPages{The International Archives of the Photogrammetry, Remote Sensing and Spatial Information Sciences}{43}{}{349--354,}
\newblock

\newblock

\PrintBackRefs{\CurrentBib}

\bibitem [\protect \citeauthoryear {%
Ren%
, He%
, Girshick%
\BCBL {}\ \BBA {} Sun%
}{%
Ren%
\ \protect \BOthers {.}}{%
{\protect \APACyear {2015}}%
}]{%
ren2015faster}
\APACinsertmetastar {%
ren2015faster}%
\begin{APACrefauthors}%
Ren, S.%
, He, K.%
, Girshick, R.%
\BCBL {} Sun, J.%
\end{APACrefauthors}%
\unskip\
\newblock
\APACrefYearMonthDay{2015}{}{}.
\newblock
{\BBOQ}\APACrefatitle {Faster r-cnn: Towards real-time object detection with region proposal networks} {Faster r-cnn: Towards real-time object detection with region proposal networks}.{\BBCQ}
\newblock
\APACjournalVolNumPages{Advances in neural information processing systems}{28}{}{,}
\newblock

\newblock

\PrintBackRefs{\CurrentBib}

\bibitem [\protect \citeauthoryear {%
Rigatti%
}{%
Rigatti%
}{%
{\protect \APACyear {2017}}%
}]{%
rigatti2017random}
\APACinsertmetastar {%
rigatti2017random}%
\begin{APACrefauthors}%
Rigatti, S.J.%
\end{APACrefauthors}%
\unskip\
\newblock
\APACrefYearMonthDay{2017}{}{}.
\newblock
{\BBOQ}\APACrefatitle {Random forest} {Random forest}.{\BBCQ}
\newblock
\APACjournalVolNumPages{Journal of Insurance Medicine}{47}{1}{31--39,}
\newblock

\newblock

\PrintBackRefs{\CurrentBib}

\bibitem [\protect \citeauthoryear {%
Szegedy%
\ \protect \BOthers {.}}{%
Szegedy%
\ \protect \BOthers {.}}{%
{\protect \APACyear {2015}}%
}]{%
szegedy2015going}
\APACinsertmetastar {%
szegedy2015going}%
\begin{APACrefauthors}%
Szegedy, C.%
, Liu, W.%
, Jia, Y.%
, Sermanet, P.%
, Reed, S.%
, Anguelov, D.%
\BDBL {}Rabinovich, A.%
\end{APACrefauthors}%
\unskip\
\newblock
\APACrefYearMonthDay{2015}{}{}.
\newblock
{\BBOQ}\APACrefatitle {Going deeper with convolutions} {Going deeper with convolutions}.{\BBCQ}
\newblock
 \APACrefbtitle {Proceedings of the IEEE conference on computer vision and pattern recognition} {Proceedings of the ieee conference on computer vision and pattern recognition}\ (\BPGS\ 1--9).
\PrintBackRefs{\CurrentBib}

\bibitem [\protect \citeauthoryear {%
Tan%
\ \BBA {} Le%
}{%
Tan%
\ \BBA {} Le%
}{%
{\protect \APACyear {2021}}%
}]{%
tan2021efficientnetv2}
\APACinsertmetastar {%
tan2021efficientnetv2}%
\begin{APACrefauthors}%
Tan, M.%
\BCBT {}\ \BBA {} Le, Q.%
\end{APACrefauthors}%
\unskip\
\newblock
\APACrefYearMonthDay{2021}{}{}.
\newblock
{\BBOQ}\APACrefatitle {Efficientnetv2: Smaller models and faster training} {Efficientnetv2: Smaller models and faster training}.{\BBCQ}
\newblock
 \APACrefbtitle {International conference on machine learning} {International conference on machine learning}\ (\BPGS\ 10096--10106).
\PrintBackRefs{\CurrentBib}

\bibitem [\protect \citeauthoryear {%
Taud%
\ \BBA {} Mas%
}{%
Taud%
\ \BBA {} Mas%
}{%
{\protect \APACyear {2018}}%
}]{%
taud2018multilayer}
\APACinsertmetastar {%
taud2018multilayer}%
\begin{APACrefauthors}%
Taud, H.%
\BCBT {}\ \BBA {} Mas, J.%
\end{APACrefauthors}%
\unskip\
\newblock
\APACrefYearMonthDay{2018}{}{}.
\newblock
{\BBOQ}\APACrefatitle {Multilayer perceptron (MLP)} {Multilayer perceptron (mlp)}.{\BBCQ}
\newblock
\APACjournalVolNumPages{Geomatic approaches for modeling land change scenarios}{}{}{451--455,}
\newblock

\newblock

\PrintBackRefs{\CurrentBib}

\bibitem [\protect \citeauthoryear {%
Tsai%
\ \BBA {} Chatterjee%
}{%
Tsai%
\ \BBA {} Chatterjee%
}{%
{\protect \APACyear {2018}}%
}]{%
tsai2018pothole}
\APACinsertmetastar {%
tsai2018pothole}%
\begin{APACrefauthors}%
Tsai, Y\BHBI C.%
\BCBT {}\ \BBA {} Chatterjee, A.%
\end{APACrefauthors}%
\unskip\
\newblock
\APACrefYearMonthDay{2018}{}{}.
\newblock
{\BBOQ}\APACrefatitle {Pothole detection and classification using 3D technology and watershed method} {Pothole detection and classification using 3d technology and watershed method}.{\BBCQ}
\newblock
\APACjournalVolNumPages{Journal of Computing in Civil Engineering}{32}{2}{04017078,}
\newblock

\newblock

\PrintBackRefs{\CurrentBib}

\bibitem [\protect \citeauthoryear {%
Zhang%
}{%
Zhang%
}{%
{\protect \APACyear {2013}}%
}]{%
zhang2013advanced}
\APACinsertmetastar {%
zhang2013advanced}%
\begin{APACrefauthors}%
Zhang, Z.%
\end{APACrefauthors}%
\unskip\
\newblock
\APACrefYear{2013}.
\unskip\
\newblock
\APACrefbtitle {Advanced stereo vision disparity calculation and obstacle analysis for intelligent vehicles} {Advanced stereo vision disparity calculation and obstacle analysis for intelligent vehicles}\ \APACtypeAddressSchool {\BUPhD}{}{}.
\unskip\
\newblock
\APACaddressSchool {}{University of Bristol}.
\PrintBackRefs{\CurrentBib}

\end{thebibliography}

\end{document}